\documentclass[runningheads]{llncs}
\usepackage[T1]{fontenc}
\usepackage{graphicx}
\usepackage{booktabs}
\usepackage{subcaption}
\usepackage{amsmath}
\usepackage{algorithm}
\usepackage{algpseudocode}
\usepackage{hyperref}
\usepackage{amssymb}
\usepackage{multirow}
\usepackage{xr}
\usepackage{cleveref}
\usepackage{array}
\usepackage{xcolor}
\usepackage[misc]{ifsym}
\newcommand{\corr}{(\Letter)}
\usepackage{mwe}

\begin{document}

\title{Learning from Stochastic Teacher Representations Using Student-Guided Knowledge Distillation}

\titlerunning{Stochastic Self Distillation}


\author{Muhammad Haseeb Aslam\inst{1} \corr \and Clara Martinez \inst{2} \and Marco Pedersoli \inst{1} \and Alessandro Lameiras Koerich \inst{1} \and Ali Etemad \inst{3} \and Eric Granger \inst{1}  }


\authorrunning{Aslam et al.}

\institute{LIVIA, Dept. of Systems Engineering, ETS Montreal, Canada \email{muhammad-haseeb.aslam.1@ens.etsmtl.ca \\ \{eric.granger, marco.pedersoli, alessandro.koerich\}@etsmtl.ca }  \and
CentraleSupélec, Universite Paris Saclay, Paris, France
\and
Aiim Lab, Queen's University, Canada.
\\ \email{ali.etemad@queensu.ca}}

\tocauthor{Muhammad Haseeb Aslam \corr, Clara Martinez, Marco Pedersoli, Alessandro Lameiras Koerich, Ali Etemad, Eric Granger}
\toctitle{Learning from Stochastic Teacher Representations Using Student-Guided Knowledge Distillation}

\maketitle              

\begin{abstract}
 Advances in self-distillation have shown that when knowledge is distilled from a teacher to a student using the same deep learning (DL) model, student performance can surpass the teacher,  particularly when the model is over-parameterized and the teacher is trained with early stopping. Alternatively, ensemble learning also improves performance, although training, storing, and deploying multiple DL models becomes impractical as the number of models grows. Even distilling a deep ensemble to a single student model or weight averaging methods first requires training of multiple teacher models and does not fully leverage the inherent stochasticity for generating and distilling diversity in DL models. These constraints are particularly prohibitive in resource-constrained or latency-sensitive applications on, e.g., wearable devices.
 This paper proposes to train only one model and generate multiple diverse teacher representations using \textit{distillation-time dropout}. However, generating these representations stochastically leads to noisy representations that are misaligned with the learned task. To overcome this problem, a novel stochastic self-distillation (SSD) training strategy is introduced for filtering and weighting teacher representation to distill from task-relevant representations only, using student-guided knowledge distillation. The student representation at each distillation step is used to guide the distillation process. 
 Experimental results\footnote{Code and supplementary available at:   \href{https://github.com/haseebaslam95/SSD}{https://github.com/haseebaslam95/SSD}} on real-world affective computing, wearable/biosignal (UCR Archive), HAR, and image classification datasets show that the proposed SSD method can outperform state-of-the-art methods without increasing the model size at both training and testing time. It incurs negligible computational complexity compared to ensemble learning and weight averaging methods.    

\end{abstract}

\section{Introduction}

Wearable technology has many applications, primarily in healthcare monitoring, such as activity and exercise tracking, sleep analysis, stress detection, and fall detection. It also includes applications like chronic disease management, personalized health insights, and human behavior and physiology research by continuously tracking metrics like heart rate, steps taken, body temperature, and movement patterns over time.  Time-series signals such as electrocardiogram (ECG), respiration rate, and other biosignals are often multi-dimensional, noisy, and collected in real time from resource-constrained devices. These signals require efficient processing methods that balance accuracy with computational efficiency. Cumbersome methods for performance boosting are less effective for this application. Knowledge distillation (KD) is typically used for transferring knowledge from a large, well-trained teacher model to a more compact student model for deployment, thereby enhancing the latter's accuracy without incurring significant computational costs \cite{hinton2015kd}. 

Self-distillation is a specialized case in KD, where the teacher and student have the same DL architecture, and the student typically surpasses the teacher's performance particularly where the  model is over-parameterized i.e., has sufficient capacity and the teacher is trained with early-stopping. This increase in performance is typically associated with the fact that, with DL models, the teacher and the student have learned separate discriminative features, and self-distillation implicitly ensembles the two models \cite{allenzhu2023}. Diversity in the feature space is a critical factor that enhances the robustness and accuracy of machine learning models. Diverse representations provide a comprehensive understanding of the input data, mitigating overfitting and improving generalization across various tasks \cite{fort2020deepensembleslosslandscape} \cite{lakshminarayanan2017simplescalablepredictiveuncertainty}. 

Approaches for ensemble learning leverage the independent training of multiple diverse models to learn more robust decision boundaries, leading to significant improvements in predictive accuracy. Despite these advantages, deploying deep ensembles introduces substantial computational and storage overhead, as each model in the ensemble requires independent training, parameter storage, and inference pipelines. These constraints are particularly prohibitive in resource-constrained embedded systems as employed in wearable applications. 

State-of-the-art (SOTA) approaches \cite{allenzhu2023} that distill diverse ensemble-based representations involve the cumbersome process of training the teacher model multiple times or utilizing complex ensemble learning methods to generate a pool of diverse teacher models for effective knowledge transfer. These methods are computationally intensive and may not fully leverage the potential of stochasticity inherent in DL models for generating diversity.

This paper introduces a KD training strategy called Stochastic Self-Distillation (SSD) to capitalize on \textit{distillation-time} dropout, thereby inducing stochasticity in a single, pre-trained teacher model. 
SSD generates multiple stochastic feature representations, effectively simulating a diverse ensemble of DL models without requiring extensive teacher re-training. This technique aligns with the principles of Monte Carlo dropout
\cite{gal2016dropoutbayesianapproximationrepresenting}. Moreover, a Student-Guided Knowledge Distillation (SGKD) is introduced to distill the most relevant knowledge (or filter out noisy representations) to the student model using student-guided attention. This mechanism allows the student to selectively focus on the most informative representations within the teacher's output space, facilitating a more efficient and targeted knowledge transfer. Subsequently, feature-level KD is employed to align the student's feature representations with the filtered and attention-weighted teacher feature representation.

The main contributions of this paper are summarized as follows. 

\noindent\textbf{(1)} We propose SSD, a novel distillation-time dropout strategy to generate diverse stochastic representations from a single, pre-trained teacher model. 

\noindent\textbf{(2)} Within SSD, a novel SGKD mechanism enables the student model to selectively distill knowledge from the most informative teacher representations. Feature-based KD is used to align the student's internal feature space with the teacher, promoting a more granular knowledge transfer. 

\noindent\textbf{(3)} Our extensive experiments on challenging affective computing benchmark datasets (Biovid Pain and StressID), biosignal/wearable datasets (from the UCR Archive), the HAR dataset, and benchmark image classification datasets (CIFAR-10 and CIFAR-100) show that our SSD training strategy allows training models that can achieve SOTA performance while maintaining computational efficiency.

\section{Related Work}

\noindent\textbf{Knowledge Distillation.} Originally introduced by ~\cite{KD-caruana} ~\cite{hinton2015kd}, the KD domain has evolved with several refinements in its application and architecture. Romero et al.~\cite{romero2015fitnetshintsdeepnets} introduced the concept of distilling from feature representations instead of logits. The idea of transferring the attention maps from the teacher model to the student model was studied by Zagoruyko and Komodakis \cite{zagoruyko2017payingattentionattentionimproving}. Relational KD proposed by Park et al.~\cite{park2019relationalknowledgedistillation} studied the benefits of utilizing structural information for more fine-grained KD. The KD domain was extended to multi-task, semi-supervised, and unsupervised learning by Lopez et al.~\cite{lopez2017multi}. KD has also been studied in multimodal systems, particularly with applications like cross-modal KD \cite{Sarkar_Etemad_2024} privileged KD \cite{PKD}, federated learning \cite{lin2021ensembledistillationrobustmodel}, are a few examples of the widespread application of KD in real-world systems.

\noindent\textbf{Deep Ensembles and Model Soups.} Ensembling methods improve predictive performance, generalization in neural networks, and uncertainty estimation. Deep ensemble is a simple yet effective technique where a simple aggregation of independently trained models harnesses the diversity, leading to better performance than each model. Lakshminarayanan et al.~\cite{lakshminarayanan2017simplescalablepredictiveuncertainty} demonstrated the effectiveness of deep ensembles for uncertainty estimation, showing that they outperform many Bayesian approaches in terms of both calibration and robustness. Deep theoretical insights on ensemble diversity were provided by Fort et al.~\cite{fort2020deepensembleslosslandscape}. Moreover, Ovadia et al.~\cite{ovadia2019} highlight the advantages of deep ensembles in handling distributional shifts, reinforcing their utility in real-world scenarios. Despite their advantages, deep ensembles are computationally expensive, requiring the training and storage of multiple models, which motivates research into alternative methods that capture similar benefits with reduced complexity. 
More recently, parameter-efficient fine-tuning techniques like low-rank adaptation (LoRA) \cite{lora} have enabled efficient fine-tuning of large models. For example, Li et al.~\cite{lora-ens} introduced Ensembles of Low-Rank Expert Adapters. However, these techniques still require i) careful adaptation of each model, and ii) storing all the models for inference.

Model soups \cite{wortsman2022modelsoupsaveragingweights}, is a technique for improving model generalization by averaging the weights of multiple fine-tuned models. Instead of selecting a single best model, model soups combine the parameters of different models fine-tuned with different hyperparameters, datasets, or random seeds, resulting in a more robust model. Two variations of the model soups were proposed: (\textit{i}) uniform soup averages the weights of all fine-tuned models equally, and (\textit{ii}) greedy soup, where models are added iteratively using a greedy approach. Model soup does not increase the model size for inference/deployment, yet it incurs significant additional train-time computational cost by fine-tuning models multiple times.  

\noindent\textbf{Self-Distillation.} This term 
has been used in the literature in two different contexts: distilling knowledge from deeper layers in a model to shallower layers of the same model's instance or through the use of an auxiliary network \cite{li-self,zhang2021self}, and knowledge distilled from a model to another instance of the model with the same architecture \cite{allenzhu2023,furlanello2018,mobahi2020}. In this work, 'self-distillation' refers to the latter. 
Furlanello et al.~\cite{furlanello2018} proposed born-again neural networks, a seminal work exploring KD using the same model for teacher and student, showing that the student can outperform the teacher. Iterative distillation from the trained student, used as a teacher for the subsequent student model, also improved performance. Dong et al.~\cite{dong2019} showed that early stopping is crucial in harnessing dark knowledge in self-distillation settings. Dark knowledge is the hidden class relationships encoded in the teacher model’s soft probability outputs, which provide more information than hard labels. This nuanced information helps the student model learn better generalization and richer representations. A direct correlation between the diversity in the teacher predictions and student performance was studied in depth by Zhang et al.~\cite{zhang2020labelsmoothing}. The authors enhanced the predictive diversity through a novel instance-specific label smoothing. 

The concept of self-distillation in a regression setting was first studied by Mohabi et al.~\cite{mobahi2020}, in which the authors provided a theoretical analysis of self-distillation where only the soft labels from the teachers were used to train the student. Multi-round self-distillation settings limit the number of basic functions that must be learned. Borup et al.~\cite{borup2021} build upon the previous analysis by including the weighted-ground truth targets in the self-distillation procedure. They show that for fixed distillation weights, the ground-truth targets lessen the sparsification and regularization effect of the self-distilled solution. 
Stanton et al.~\cite{stanton2021} studied the paradigm of KD through the lens of fidelity. Their key takeaway regarding KD and ensembles was that the highest-fidelity student is the best calibrated, even when it is not the most accurate. The closest work to SSD was proposed by Allen-Zhu and Li \cite{allenzhu2023}, who explored the concept of self-distillation in conjunction with the \textit{multi-view} structure of the input data. In this case, the student model was trained on the ground truth labels with additional supervision from the ensemble of multiple teachers' soft labels. Multiple teachers were trained with random seed initialization. 

In contrast to these methods, our proposed SSD training strategy obviates the need for such data augmentations or random seed initialization to generate diversity in the teacher space. SSD operates in the feature space, using dropout as a tool to introduce diversity and student-guided attention to distill relevant information for the student. Consequently, SSD requires significantly less complexity for training when compared to traditional ensemble learning and weight-averaging methods, and without increasing model size at deployment time.

\section{Proposed Method}

\noindent \textbf{Notation:} Let $\mathcal{T}$ be a teacher model with model parameters $\mathcal{\theta^T}$ and $\mathcal{S}$ be a student model with parameters $\mathcal{\theta^S}$. For given inputs $\mathcal{X}=[x_1, x_2,\dots,x_{m}]$, we obtain the feature vectors $f^\mathcal{T}=\mathcal{T}(\mathcal{X;\theta^\mathcal{T}})$ $\in 
\mathbb{R}^{d\times m}$  and $f^\mathcal{S}=\mathcal{S}(\mathcal{X;\theta^S})$ $\in 
\mathbb{R}^{d\times m}$, where $d$ is the dimension of the feature vector and $m$ is the number of input samples. Let $\mathcal{F}^\mathcal{T}(x)=[f^\mathcal{T}_1(x),f^\mathcal{T}_2(x),\dots, f^\mathcal{T}_n(x)]$ represents the multiple stochastic teacher representations generated through $n$ forward passes through $\mathcal{T}(\mathcal{X;\theta^T})$ for the same input sample $x \in \mathcal{X}$.

\noindent \textbf{Problem Definition:} Given a trained teacher model $\mathcal{T}$, the challenge is to effectively transfer its knowledge to a student model $\mathcal{S}$ such that its generalization performance is maximized. Standard self-distillation techniques often treat the teacher's outputs as deterministic, failing to exploit the inherent stochasticity that can provide richer and more diverse information. On the other hand, stochastically obtaining the teacher representations introduces diversity but at the cost of generating noisy representations. The problem, therefore, is to design a KD framework that leverages the variability in the teacher's representations, generated through stochastic mechanisms like dropout, while ensuring that the student learns task-relevant information in a computationally efficient manner. The main aim of our paper is to filter out the noisy representations from $\mathcal{F}^\mathcal{T}(x)$ 
and obtain weighted teacher representation $\hat{f}^{\mathcal{T}}(x)$ to selectively distill from the relevant teacher representations.

\begin{figure*}[!h]
 \centering
  \includegraphics[width=0.90\linewidth]{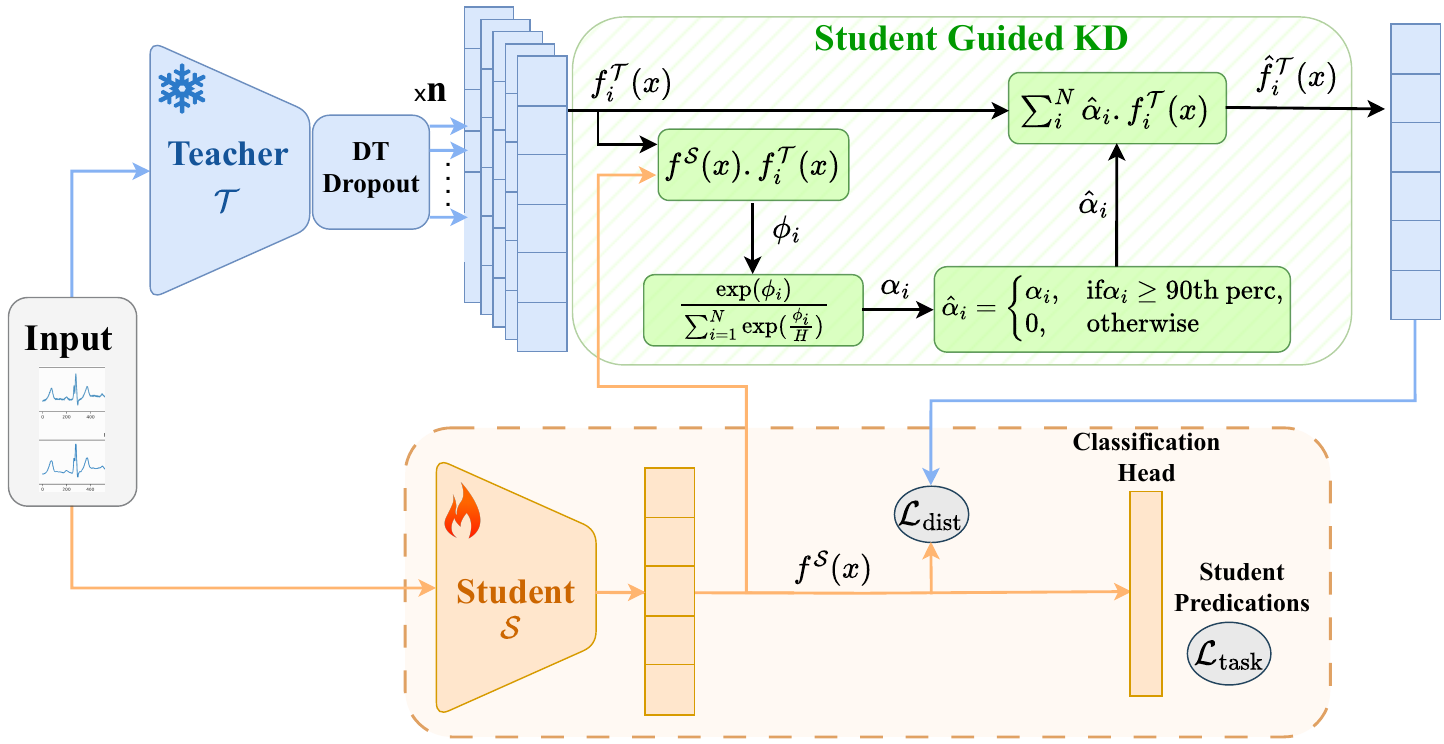}
  \caption{Illustration of the proposed SSD training strategy. The teacher $\mathcal{T}$ is trained, and its parameters are frozen except for the dropout layers in the student training stage. In SGKD, for each input $x \in \mathcal{X}$, $n$ stochastic teacher representations $f^\mathcal{T}_{i}(x)$, $i=1, 2, ..., n$, and one student representation is obtained $f^\mathcal{S}$, which are fed to the SSD module that outputs $\hat{f}^\mathcal{T}(x)$. Feature-based KD is then applied on $\hat{f}^\mathcal{T}(x)$ and $f^\mathcal{S}(x)$, with addition of $\mathcal{L}_\text{task}$, and the student model parameters $\theta_\mathcal{S}$ are updated. The part inside the dashed orange block is kept at inference.}
  \label{fig:ssd-main}
\end{figure*}

\subsection{Stochastic Self-Distillation}
The proposed SSD training strategy generates multiple diverse representations per sample and uses the current student representation as a reference to rank, select, and weigh (using student-guided attention) the teacher representations before distilling. Further, SSD enforces the attention weights of the meaningful representations to be spread out through temperature scaling, this implicitly models the feature ensemble to harness diversity. The student representation guides each distillation step because it is initialized with the same weights as the main trained teacher. Fig.~\ref{fig:ssd-main} illustrates the proposed SSD method. The remainder of this section provides details on the SSD training strategy. 

\noindent\textbf{Teacher Training.}
 The first step is to train the teacher model and get trained teacher parameters $\theta^\mathcal{T'}$. This step is needed for two purposes: i) because the trained teacher model is used to generate the stochastic teacher representations $\mathcal{F}^\mathcal{T}(x)$, and ii) because these weights are also used to initialize the student model parameters in the student training step.
 
 \noindent\textbf{Student Parameters Initialization.} After the teacher model is trained, the student model parameters are initialized with the trained teacher weights. This initialization is also crucial in the proposed training strategy. The proposed method relies on student guidance to obtain the attended teacher feature vector $\hat{f}^\mathcal{T}(x)$. The initialization of the student parameters with trained teacher weights lets the student serve as authority to weigh the teacher representations. As mentioned earlier, the stochastic nature of $\theta^\mathcal{T'}$, necessitates that some of the generated representations would be misaligned with the learned task-specific representation and hence would act as noise for the student model. This phenomenon is studied in detail in Section \ref{sub:stu_weight}

\subsection{Student-Guided Knowledge Distillation}

Traditionally, in KD methods, the teacher representation(s) are informative and serve as additional supervision for the student model. However, if the teacher representations are generated through a stochastic process, they are not aligned with the learned class boundary and can be noise for the student model. Therefore, we use the current student representation as authority to select representations aligned with the learned class boundary. The student representation at each distillation step can be used as the guiding mechanism because the student model is initialized with the learned weights from the teacher network. This initialization allows the student to make use of its intermediate representation $f^\mathcal{S}(x)$ as an anchor to rank the teacher representation $f^\mathcal{T}_{i}(x)\in\mathcal{F}^\mathcal{T}$.


To guide the distillation process using the current student representation, we first calculate the dot product ($\phi_{i}$) between the current student representation $f^\mathcal{S}(x)$ and each teacher representation $f^\mathcal{T}_{i}(x)\in\mathcal{F}^\mathcal{T}(x)$ 
and compute attention weights using: 
\begin{equation}
     \alpha_{i} = \frac{\exp(\phi_{i}/h)}{\sum_{i=1}^{N} \exp(\phi_{i}/h)}
\label{eq:attweight}
\end{equation}
where $N$ is the number of teacher representations and $h$ is a regularization factor used to smooth the attention weights to ensure that the attended teacher feature vector $\hat{f}^{\mathcal{T}}(x)$ is not heavily influenced only by a single teacher representation. This regularization step is crucial since it dictates the attention weights for teacher representations. Since the way these teacher representations are ranked is through dot product between current student representation $f^{\mathcal{\mathcal{S}}}(x)$ and each of the stochastic teacher representation $f^{\mathcal{T}}_{i}(x)$, the value $\phi_{i}$ naturally would be the highest for the teacher representation that is the most similar to the student representation. This renders the entire framework ineffective because  $\hat{f}^{\mathcal{T}}(x)$ becomes overly similar to  $f^{\mathcal{\mathcal{S}}}(x)$.

The SSD method relies on selecting teacher representations that would mimic an ensemble of independently trained teacher models. In other words, it masks out the representations that are too different from the current student representations. A direct way of selecting such representations would be to use the top-$k$ strategy at each distillation step. Although this strategy can work, it  relies on \textit{k} a hyperparameter that is agnostic to the distribution of $f^\mathcal{T}_{i}(x)\in\mathcal{F}^\mathcal{T}$ at each distillation step. To avoid a manual selection of meaningful representations using a top-$k$ strategy, $\alpha_{i}$ is masked for all indices falling outside of the $\epsilon$-th percentile, denoted by $\hat{\alpha}_{i}$. 
\begin{equation}
     \hat{\alpha}_i = \begin{cases} 
    \alpha_i, & \text{if } \alpha_i \geq \text{$\epsilon$}, \\
    0, & \text{otherwise.}
    \end{cases} 
\label{eq:percentile}
\end{equation}
\noindent where $\epsilon \in [0,100]$ is the threshold value for masking. Section \ref{subsec:abl} provides a more detailed discussion. After obtaining the regularized attention weights $\hat{\alpha}_i$, the original teacher feature representations $f^{\mathcal{T}}_{i}(x)$ are weighed using $\hat{\alpha}_i$ to obtain the attended teacher feature vector $\hat{f}^{\mathcal{T}}(x)$ as: 
\begin{equation}
    \hat{f}^{\mathcal{T}}(x) = \sum_{i=1}^{N} \hat{\alpha}_i\cdot f^{\mathcal{T}}_{i}(x)
\label{eq:weighteacher}
\end{equation}
The attended feature vector $\hat{f}^{\mathcal{T}}(x)$ is used to distill the information using the mean squared error loss:
\begin{equation}
\mathcal{L}_{\text{dist}} =\frac{1}{d} \sum_{j=1}^{d} \left( f^{\mathcal{S}}(x_j) - \hat{f}^{\mathcal{T}}(x_j) \right)^2
\label{eq:dist}
\end{equation}
The total loss for the student is shown in Eq.~\ref{eq:losstotal},  
\begin{equation}
\mathcal{L}_{\text{total}} = \mathcal{L}_{\text{task}} + \lambda\mathcal{L}_{\text{dist}}
\label{eq:losstotal}
\end{equation}
where $\lambda$ is the weighting parameter for the distillation loss. The method also allows for the additional constraint for logit-level distillation and can be added to $\mathcal{L}_{\text{total}}$. 
For the unsupervised/contrastive loss-based methods, we use the intermediate teacher representations to obtain the $\hat{f}^{\mathcal{T}}(x)$. In case of augmented views of the input sample, we average the distillation loss from both views. The pseudo-code for the SSD training procedure is shown in Algorithm \ref{alg:ssd}.


\label{app:sub_algo}
\begin{algorithm}[!h]
\caption{ - SSD Training Procedure.}
\begin{algorithmic}[1]
\Require Teacher model $\mathcal{T}$ with parameters $\theta^{\mathcal{T}}$, student model $\mathcal{S}$ with parameters $\theta^{\mathcal{S}}$, input samples $\mathcal{X}=[x_1,x_2,\dots,x_{m}]$, weighting parameter $\lambda$.
\Ensure Trained student model $\mathcal{S}$.

\State Extract teacher feature vectors $f^{\mathcal{T}}$=$\mathcal{T}(\mathcal{X}; \theta^{\mathcal{T}})$ $\in \mathbb{R}^{d \times m}$.
\State Extract student feature vectors $f^{\mathcal{S}}$=$\mathcal{S}(\mathcal{X}; \theta^{\mathcal{S}}) \in \mathbb{R}^{d \times m}$.

\For{each input sample $x \in \mathcal{X}$}
    \State Generate $n$ stochastic teacher representations $\mathcal{F}^{\mathcal{T}}_i(x) = [f^{\mathcal{T}}_1(x),\dots,f^{\mathcal{T}}_n(x)]$.

    \State Compute \textbf{dot products} $\phi_i = f^{\mathcal{S}}(x) \cdot f^{\mathcal{T}}_i(x)$ for $i = 1, 2, \dots, n$. 
    \State Compute \textbf{attention weights}: $\alpha_i$ for $i = 1, 2, \dots, n$ using Eq.~\eqref{eq:attweight}
    
    \State \textbf{Mask }$\alpha_i$ for all indices outside the $\epsilon$-th percentile using Eq.~\eqref{eq:percentile}

    \State Compute \textbf{attended teacher feature vector}: $\hat{f}^{\mathcal{T}}(x)$ using Eq.~\eqref{eq:weighteacher}

    \State Compute distillation loss $\mathcal{L}_{\text{dist}}$ using Eq.~\eqref{eq:dist}
\EndFor

\State Compute total loss $\mathcal{L}_{\text{total}}$ using Eq.~\eqref{eq:losstotal}
\State Update parameters $\theta^{\mathcal{S}}$ using backpropagation
\end{algorithmic}
\label{alg:ssd}
\end{algorithm}

\section{Results and Discussion}

\subsection{Experiment Setup}

SSD is validated on: (i) real-world affective computing datasets: the Biovid Heat Pain Database \cite{biovid-ds} and StressID dataset \cite{stressid2023}, (ii) wearable and biosignal datasets from the UCR Archive \cite{dau2019ucrtimeseriesarchive}, (iii) Human Activity Recognition (HAR) dataset set from the UCI Archive \cite{Har-ds}, and (iv) on benchmark image classification datasets. Appendix A.1 provides details on these datasets, while Appendix A.3 provides implementation details and results on the CIFAR-10 and CIFAR-100 datasets.  

\noindent\textbf{Biovid Heat Pain Database.} For Biovid, we use the EDA modality and LOSO cross-validation. The proposed method is tested with a SOTA method on the dataset using the Pain Attention Net \cite{Lu_pan2023}, which is a transformer-based physiological signal classification network comprised of a multi-scale convolutional network, as SE residual network, and a transformer encoder block. The batch size used for teacher training was 128, and the network was optimized using the Adam optimizer with a learning rate of $0.001$. The network was trained over 100 epochs with early stopping. The total number of folds was 87, corresponding to the total number of subjects. For student training, we keep the same setting as the teacher. We manually activate dropout layers with a $p$-value of $0.2$ while keeping the teacher model in inference mode. The total number of repetitions for generating diverse teacher representations was 30.

\noindent\textbf{StressID Dataset.} For the StressID dataset, use the EDA and RR modalities and apply feature concatenation to fuse the backbone representations. The EDA backbone was Pain Attention Net \cite{Lu_pan2023}, and the RR backbone was a 1D CNN with three 1D conv layers with 16, 32, and 64 channels, respectively, with a kernel size of 5, and stride equal to 1, followed by three batch normalization layers. Following the original dataset authors \cite{stressid2023}, we apply an $80-20$ split for the train and test set and further divide the train data and keep $20\%$ of that for model selection. The batch size used for both teacher and student training was 128, with the learning rate of $0.001$ using the Adam optimizer. The total number of repetitions was 30. The dropout layer was activated before the feature concatenation module with a $p$-value of $0.2$.

\noindent\textbf{UCR Archive.} For the datasets in the UCR Archive, the proposed method was applied to two SOTA techniques -- TS2Vec and SoftCLT -- for unsupervised time-series representation learning. We follow the same experimental methodology proposed in TS2Vec \cite{yue2022ts2vecuniversalrepresentationtime} and SoftCLT \cite{lee2024soft}. For the TS2Vec method, the number of stochastic teacher representations was 15, with a teacher dropout rate $p$ of 0.2, and the student dropout rate was set to 0.1. The value of $H$ was 5. For loss weighting, $\lambda$ was set to 0.2. For softCLT, all the parameters were kept the same as those for TS2Vec except for the value of $H$, which was set to 15. 

\noindent\textbf{Computing infrastructure.} All experiments were performed on the NVIDIA A100-SXM4-40GB GPUs with the $\epsilon$ value of 90.

\subsection{Comparison Against State-of-the-Art Methods}


\noindent\textbf{Affective Computing Datasets.} Table \ref{tab:bio_sota} reports the results of SSD and SOTA methods on Biovid. SSD improves accuracy by $2.5\%$ over the selected baseline (PAN without SSD) and $1.7\%$ over the current SOTA. Specifically, accuracy on Biovid increases from $84.59\%$ (using the EDA modality without SSD) to $86.90\%$ with the proposed SSD method.

\begin{table*}[!t]
\centering
\small
\caption{Accuracy of SSD against state-of-the-art methods on Biovid data. Baseline results were obtained using the network architecture without applying the proposed SSD method (without distillation).}
\begin{tabular}{ll|c|c|c}
\hline
\multicolumn{2}{c|}{\textbf{Method}} & \textbf{Modality}     & \textbf{CV Scheme} & \textbf{Accuracy } \\ \hline \hline
Werner et al.   \cite{werner2014automatic}    & ICPR 2014           & Physio + Vision & 5-fold             & 0.8060                   \\ 
Werner et al.   \cite{werner2016automatic}   &   IEEE TAC 2016           & Video                 & LOSO               & 0.7240                   \\ 
Kachele et al. \cite{kachele2016methods}     & IEEE IJSTSP 2016            & EDA, ECG, EMG         & LOSO               & 0.8273                  \\ 
Lopez et al.      \cite{lopez2017multi}          & ACII 2017     & EDA, ECG              & 10-fold            & 0.8275                  \\ 
Lopez et al. \cite{lopez2018continuous}     & EMBC 2018               & EDA                   & LOSO               & 0.7421                  \\ 
Thiam et al.  \cite{thiam2019exploring}        & Sensors 2019          & EDA                   & LOSO               & 0.8457                  \\ 
Wang et al. \cite{wang2020hybrid}         &  EMBC 2020          & EDA, ECG, EMG         & LOSO               & 0.8330                   \\ 
Pouromran et al. \cite{pouromran2021exploration}  & PLoSONE 2021             & EDA                   & LOSO               & 0.8330                   \\ 
Thiam et al \cite{thiam2021multi}             & Frontiers 2021       & EDA, ECG, EMG         & LOSO               & 0.8425                  \\ 
Shi et al \cite{shi-2022}          &   ICOST 2022        & EDA         & LOSO               & 0.8523                  \\ 
Ji et al \cite{ji_2023}          & ACM SAC 2023           & EDA    & LOSO               & 0.8040                  \\ 
Jiang et al \cite{JIANG2024121082}         & ESWA 2024             & EDA    & LOSO               & 0.8458                  \\ 
 \hline 
Baseline (w/o SSD)        &    --      & EDA                   & LOSO               &  0.8459          \\
\textbf{SSD (ours)}       &    --      & \textbf{EDA}          & \textbf{LOSO}      & \textbf{0.8690}          \\ \hline
\end{tabular}
\label{tab:bio_sota}
\end{table*}
Table \ref{tab:sota_sid} compares the performance of the proposed method with SOTA on the StressID dataset. We achieve 0.74$\pm$0.02 for the F1-score and 0.74$\pm$0.03 accuracy without applying SSD. The proposed method improves 3$\%$ for the F1-score and 4$\%$ in accuracy for the binary classification task over the SOTA. This increase in predictive performance shows that the proposed method can perform well on real-world time-series dataset tasks by effectively harnessing task-relevant diversity from stochastic teacher representations.

\begin{table*}[!h]
\centering
\small
\caption{Performance of the SSD against state-of-the-art methods on the StressID dataset. (MM: Multimodal, NR: Not Reported.)}
\begin{tabular}{l|c|cc|cc}
\hline 
\multicolumn{1}{c|}{\multirow{2}{*}{\centering \textbf{Method}}} & \multicolumn{1}{c|}{\multirow{2}{*}{\centering \textbf{Modality}}} & \multicolumn{2}{c|}{\textbf{2-class problem}}                                         & \multicolumn{2}{c}{\textbf{3-class problem}}                                         \\ \cline{3-6}
                                     &                                        & \multicolumn{1}{|c|}{\textbf{F1-score}} & \textbf{Accuracy}                    & \multicolumn{1}{c|}{\textbf{F1-score}} & \textbf{Accuracy}                   \\ \hline  \hline
 HC  + RF            & \multicolumn{1}{c|}{Physio}            & \multicolumn{1}{c|}{0.73 $\pm$ 0.02}   & 0.72 $\pm$ 0.03                      & \multicolumn{1}{c|}{0.55 $\pm$ 0.04}   & 0.56 $\pm$ 0.03                     \\ 
 HC  + SVM             & \multicolumn{1}{c|}{Physio}            & \multicolumn{1}{c|}{0.71 $\pm$ 0.02}   & 0.71 $\pm$ 0.02                      & \multicolumn{1}{c|}{0.59 $\pm$ 0.04}   & 0.59 $\pm$ 0.03                     \\ 
 HC  + MLP            & \multicolumn{1}{c|}{Physio}            & \multicolumn{1}{c|}{0.70 $\pm$ 0.03}   & 0.70 $\pm$ 0.03                      & \multicolumn{1}{c|}{0.54 $\pm$ 0.04}   & 0.53 $\pm$ 0.04                     \\ 
AUs + kNN                            & \multicolumn{1}{c|}{Vision}            & \multicolumn{1}{c|}{0.70 $\pm$ 0.04}   & 0.69 $\pm$ 0.04                      & \multicolumn{1}{c|}{0.54 $\pm$ 0.05}   & 0.53 $\pm$ 0.05                     \\ 
AUs + SVM                            & \multicolumn{1}{c|}{Vision}            & \multicolumn{1}{c|}{0.69 $\pm$ 0.04}   & 0.69 $\pm$ 0.04                      & \multicolumn{1}{c|}{0.55 $\pm$ 0.05}   & 0.54 $\pm$ 0.04                     \\ 
AUs + MLP                             & \multicolumn{1}{c|}{Vision}            & \multicolumn{1}{c|}{0.70 $\pm$ 0.03}   & 0.70 $\pm$ 0.03                      & \multicolumn{1}{c|}{0.55 $\pm$ 0.03}   & 0.55 $\pm$ 0.03                     \\ 
 HC  + kNN               & \multicolumn{1}{c|}{Audio}             & \multicolumn{1}{c|}{0.67 $\pm$ 0.06}   & 0.60 $\pm$ 0.05                      & \multicolumn{1}{c|}{0.53 $\pm$ 0.04}   & 0.52 $\pm$ 0.04                     \\ 
 HC  + SVM              & \multicolumn{1}{c|}{Audio}             & \multicolumn{1}{c|}{0.61 $\pm$ 0.06}   & 0.54 $\pm$ 0.03                      & \multicolumn{1}{c|}{0.53 $\pm$ 0.08}   & 0.48 $\pm$ 0.04                     \\ 
wav2vec 2.0                     & \multicolumn{1}{c|}{Audio}             & \multicolumn{1}{c|}{0.70 $\pm$ 0.02}   & \multicolumn{1}{c|}{0.66 $\pm$ 0.03} & \multicolumn{1}{c|}{0.56 $\pm$ 0.04}   & \multicolumn{1}{c}{0.52 $\pm$ 0.04} \\ 
Mordacq et al.                  & \multicolumn{1}{c|}{MM}             & \multicolumn{1}{c|}{0.69 }   & \multicolumn{1}{c|}{0.76} & \multicolumn{1}{c|}{NR}   & \multicolumn{1}{c}{NR} \\ 
\hline
Baseline (w/o SSD)                           & \multicolumn{1}{c|}{Physio}            & \multicolumn{1}{c|}{0.74 $\pm$ 0.02}   & \multicolumn{1}{c|}{0.74 $\pm$ 0.03} & \multicolumn{1}{c|}{0.61 $\pm$ 0.01}   & \multicolumn{1}{c}{0.59 $\pm$ 0.01} \\ 
\textbf{SSD (ours) }                          & \multicolumn{1}{c|}{\textbf{Physio}}            & \multicolumn{1}{c|}{\textbf{0.77 $\pm$ 0.03}}   & \multicolumn{1}{c|}{\textbf{0.77 $\pm$ 0.03}} & \multicolumn{1}{c|}{\textbf{0.63 $\pm$ 0.02}}   & \multicolumn{1}{c}{\textbf{0.60 $\pm$ 0.02}} \\ \hline
\end{tabular}
\label{tab:sota_sid}
\end{table*}

\noindent\textbf{Time-Series Datasets (UCR Archive).}
Table \ref{tab:sota-wearablebio} shows the performance of SSD against the TS2Vec baseline on 12 wearable/biosignal datasets from the UCR Archive. The student model achieves an average accuracy score of $0.8441$.

\begin{table*}[!h]
\centering
\caption{Accuracy of SSD applied to TS2Vec and SoftCLT baselines for the wearable biosignal datasets in the UCR Archive.}
\small
\begin{tabular}{l|c|c|c|c}
\hline
     \multirow{2}{*}{\centering \textbf{Dataset}}    & TS2Vec \cite{yue2022ts2vecuniversalrepresentationtime} & \textbf{TS2Vec +}  & SoftCLT \cite{lee2024soft} & \textbf{SoftCLT +} \\
\multicolumn{1}{c|}{}       & AAAI '22  & \textbf{SSD (Ours)} & ICLR '24  & \textbf{SSD (Ours)}  \\ \hline \hline
ECG200                     & 0.9000          & \textbf{0.9100}      & 0.8800           & \textbf{0.9300}      \\
ECG5000                    & 0.9348          & \textbf{0.9411}      & 0.9400           & \textbf{0.9413}      \\
TwoLeadECG                 & 0.9789          & \textbf{0.9877}      & 0.9762           & \textbf{0.9798}      \\
NIFetalECGThorax1 & 0.9277          & \textbf{0.9318}      & 0.9201           & \textbf{0.9394}      \\
NIFetalECGThorax2 & \textbf{0.9389} & 0.9343               & 0.9435           & \textbf{0.9480}      \\
Chinatown                  & \textbf{0.9737} & 0.9708               & \textbf{0.9737}  & 0.9708               \\
UWaveGestureLibraryX       & 0.7995          & \textbf{0.8079}      & 0.8001           & \textbf{0.8143}      \\
UWaveGestureLibraryY       & 0.7152          & \textbf{0.7317}      & 0.7169           & \textbf{0.7266}      \\
UWaveGestureLibraryZ       & 0.7624          & \textbf{0.7660}      & 0.7674           & \textbf{0.7682}      \\
MedicalImages              & \textbf{0.8092} & 0.8078               & \textbf{0.8171}  & 0.7710               \\
DodgerLoopDay              & 0.5125          & \textbf{0.5250}      & 0.5500           & 0.5500               \\
DodgerLoopGame             & 0.7826          & \textbf{0.8405}      & 0.8260           & \textbf{0.8695}      \\ \hline
Total                      & 0.8350          & \textbf{0.8441}      & 0.8426           & \textbf{0.8508}               \\ \hline
\end{tabular}
\label{tab:sota-wearablebio}
\end{table*}

\noindent\textbf{Comparison with Traditional Ensembles and Model Soups}
Table \ref{tab:ens-har} compares the performance of SSD against traditional ensembles and weight-averaging methods. For simplicity in experimentation and to make sure the results are not biased by the internal mechanisms of architecture, this comparison is performed on bare-bones 1D CNN architecture, which serves as the teacher network for SSD and is also used in the ensembles as well as weight-averaging results.

\begin{table}[!h]
\centering
\caption{Accuracy of the SSD against state-of-the-art methods with a 1D CNN and traditional ensembles on the HAR dataset.}
\begin{tabular}{l|c}
\hline
\multicolumn{1}{c|}{\textbf{Model}}  & \multicolumn{1}{c}{\textbf{Accuracy}} \\ \hline \hline
1D CNN (Baseline) & 0.9002   \\ 
Ensemble Majority Vote (25 Models)   & 0.9135   \\
Ensemble Average (25 Models)   & 0.9128   \\
Model Soup (10 Models) & 0.9101   \\
Model Soup (25 Models)  & 0.9183   \\
SSD (Student)   & 0.9182   \\ \hline
\end{tabular}

\label{tab:ens-har}
\end{table}

\begin{figure}[!h]
    \centering
    \includegraphics[width=0.90\linewidth]{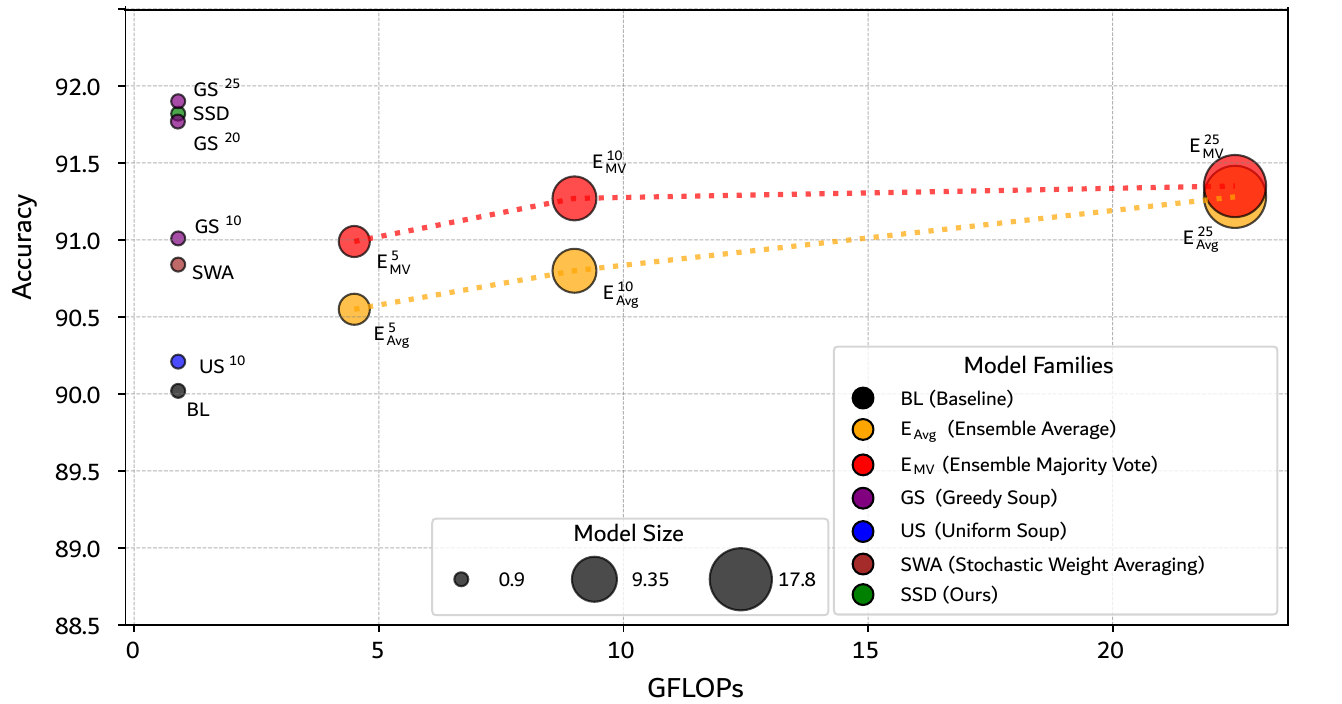}
    \caption{Comparison of the SSD method with baseline (BL), traditional ensembles with majority vote ($\text{E}_{\text{MV}}$) and average ($\text{E}_\text{Avg}$), uniform soup (US) and greed soups (GS) on HAR dataset in terms of accuracy and model size at inference.}
    \label{fig:comp-perf-flops}
\end{figure}

SSD aims to minimize the space and computational complexity both during training and testing. Fig.~\ref{fig:comp-perf-flops} compares the performance gain in terms of model size at inference. We compare traditional ensembles (majority voting and averaging), stochastic weight averaging, uniform soup (uniform weight averaging), and greedy soup (a greedy approach for weight averaging). 
The marker size in Fig.~\ref{fig:comp-perf-flops} denotes the model size at inference; since traditional ensembles require storing all of the trained models for inference, it becomes impractical to deploy for inference on wearable devices. It can be observed from Fig.~\ref{fig:comp-perf-flops} that both majority vote and averaging-based ensembles increase the model performance, but the model size proportionally increases; essentially, for an ensemble model with 25 models, it would become \textbf{25$\times$} the size of the baseline model. On the other hand, model soups do not increase the model size at inference but are still computationally expensive in terms of train-time FLOPs as shown in Appendix B. The total number of FLOPs increases from $\approx$ \textbf{0.87 G-FLOPs} to \textbf{$\approx$ 21.8 G-FLOPs}. In contrast, the proposed method achieves comparable performance to the traditional approaches, i.e., 1.8\% increase in the accuracy over the baseline, while keeping the model size the same as the baseline model, and the train-time computation is significantly less since SSD requires the model to be trained twice, once in the teacher training process and second for student training.

\subsection{Ablations}
\label{subsec:abl}
\noindent\textbf{Number of Stochastic Representations.}
As discussed before, meaningful diversity in the teacher space is crucial for the superior performance of an ensemble and, by extension, also crucial in SSD, since it also implicitly ensembles teacher representations. The number of repetitions dictates how diverse $f^\mathcal{T}_{i}(x)$ is. We evaluated with different settings and reported the results in Table \ref{tab:abb-nr}. Distill All - the first two rows show the results without applying SSD and learning from all stochastic representations. This could also be seen as an alternative way of teaching students to drop out. Rows 3-6 show results with an increasing number of total repetitions and selected representations. As the total number of representations becomes too large, the performance drops even when applying SSD. This leads us to believe that when you select a more significant number of representations, the noisy representations bypass through the filtering mechanism and become part of the distillation process. This problem also indicates a simple top-$k$ selection is not the best strategy in this case. Hence, dynamic selection is applied based on $\epsilon$-th percentile thresholding.

\begin{table}[t]
\centering
\small
\caption{Comparison of schemes for selecting teacher representations on the StressID dataset. (DS: Dynamic selection. NA: Not available.)}
\begin{tabular}{c|c|c|c|c}
\hline
\textbf{Scheme}        & \textbf{\# Representations}            & \textbf{\# Selected} & \textbf{F1-Score }       & \textbf{Accuracy}        \\ \hline \hline
Distill All  & 10 &  All  & 0.74 ±   0.02   & 0.74 ± 0.02     \\
Distill All  & 30 &  All  & $\downarrow$ 0.72 ±   0.02   & 0.72 ± 0.02     \\ 
\hline
  top-$k$    & 10              & 3        & $\uparrow$ 0.75 ±   0.02   & 0.74 ± 0.02     \\
  top-$k$    & 20              & 10       & $\uparrow$ 0.76 ±   0.03   & 0.76 ± 0.02     \\
  top-$k$    & 30              & 15       & $\uparrow$ 0.76 ±   0.04   & 0.76 ±   0.04   \\
  top-$k$    & 50              & 30       & $\downarrow$ 0.72 ±   0.02   & 0.73 ±   0.02   \\
  \hline
 \textbf{DS} & \textbf{30}              & \textbf{NA}        & 
$\uparrow$ \textbf{0.77 ±   0.03} & \textbf{0.77 ±   0.03} \\ 
 \textbf{DS} & \textbf{50}              & \textbf{NA}        & 
$\uparrow$ \textbf{0.77 ±   0.03} & \textbf{0.77 ±   0.04} \\ \hline
\end{tabular}

\label{tab:abb-nr}
\end{table}

\begin{figure*}[h]
    \centering
\includegraphics[width=1\linewidth]{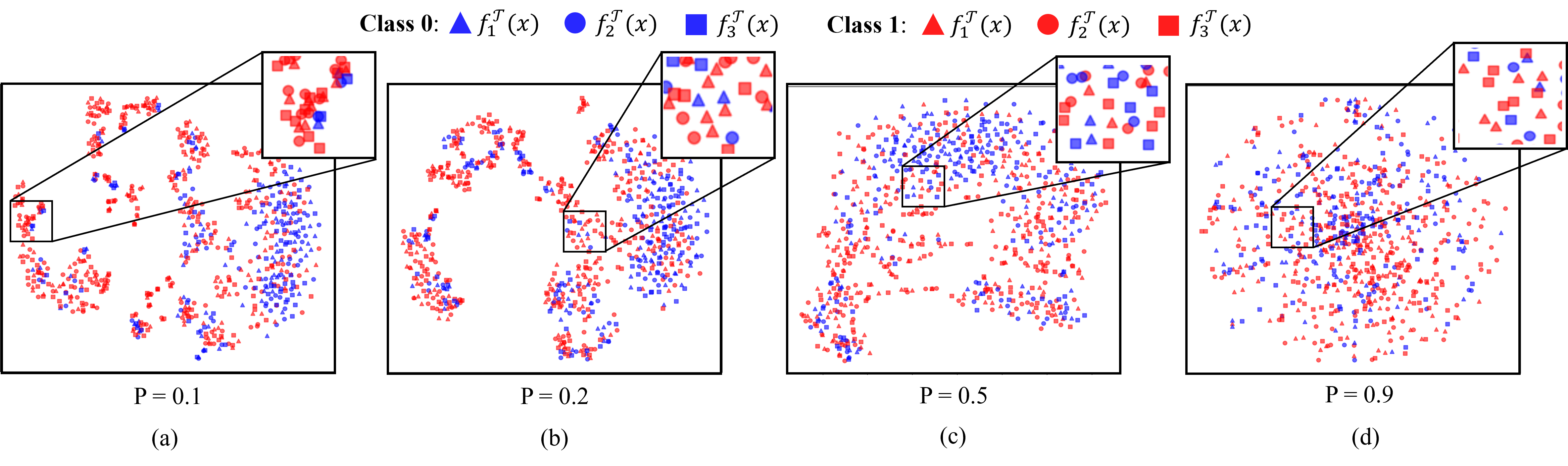}
    \caption{t-SNE plots of three teacher representations $f^{\mathcal{T}}_1(x)$ $[$triangle$]$, $f^{\mathcal{T}}_2(x)$ $[$circle$]$, $f^{\mathcal{T}}_3(x)$ $[$square$]$ over various dropout rates $P$ showing the effect of the stochastic representations on the learned feature space.}
    \label{fig:drr-abb}
\end{figure*}

\noindent\textbf{Dropout Rate} The extent of diversity in the teacher space is directly related to the distillation-time dropout rate. To study how the probability value of each neuron to be deactivated affects the teacher representation space, we plot and compare the t-SNE plots with different dropout rates. Figs.~\ref{fig:drr-abb}(a)-(d) are plotted with dropout rates $0.1$, $0.2$, $0.5$ and $0.9$ respectively. It is observed that for smaller dropout rates (Fig.~\ref{fig:drr-abb}(a)), the teacher can maintain its discriminative ability; however, the three teacher representations $f^{\mathcal{T}}_1(x)$ (triangle), $f^{\mathcal{T}}_2(x)$ (circle), $f^{\mathcal{T}}_3(x)$ (square) mostly overlap each other, effectively meaning there is not enough diversity in the teacher space. For dropout rate $0.2$ (Fig.~\ref{fig:drr-abb}(b)), the three representations are diverse while maintaining the original structure, which shows that the teacher space has become diverse while maintaining the discriminative ability. In Fig.~\ref{fig:drr-abb}(c), the three representations are adequately spaced, but the model loses its discriminative ability.

\begin{figure*}[!h]
    \centering
\includegraphics[width=0.90\linewidth]{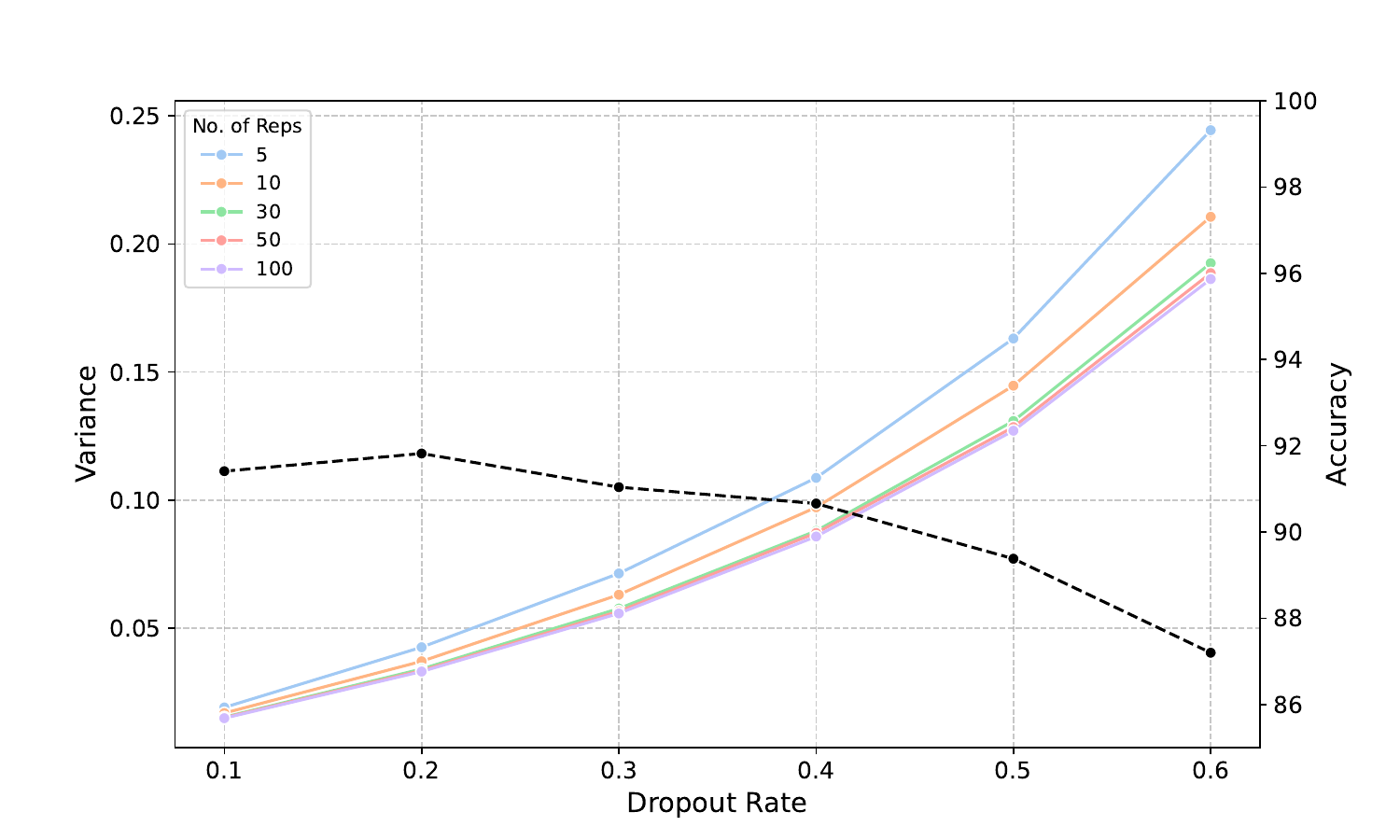}
    \caption{Effect of various dropout rates on the performance and variance quantification among teacher representations. Accuracy (black/dashed line), and variance for various no. of reps (colored/solid lines) are plotted against dropout rates on the HAR dataset}
    \label{fig:tv_tdo_acc}
\end{figure*}

To further analyze the impact dropout rate on both variance and overall performance, we conducted an ablation study on the HAR dataset. Fig.~\ref{fig:tv_tdo_acc} shows the effect of dropout rates on the student performance (dashed/black line) and the variance across teacher representations (colored/solid lines). Variance increases with the teacher dropout rate. For each dropout rate, the highest variance is observed for the lowest number of repetitions. Conversely, the lowest variance for each dropout rate is observed with the highest number of reps. Results suggest that when the number of forward passes is greater, the overall variance decreases. It may also indicate a more accurate approximation of the true variance because it is calculated with more samples. In the latter case, it can be observed that for lower dropout rates, the variances across different number of repetitions are more accurate approximation of the true variance. In terms of performance, the model peaks at a dropout rate of 0.2, and the performance starts deteriorating beyond a dropout rate of 0.5. This phenomenon can be further explained from Fig.~\ref{fig:drr-abb} where the t-SNE plots show the model starting to lose its discriminative ability at higher dropout rates.    

\begin{figure}[h]
    \centering
    \begin{subfigure}{0.48\linewidth}
        \centering
        \includegraphics[width=1\linewidth]{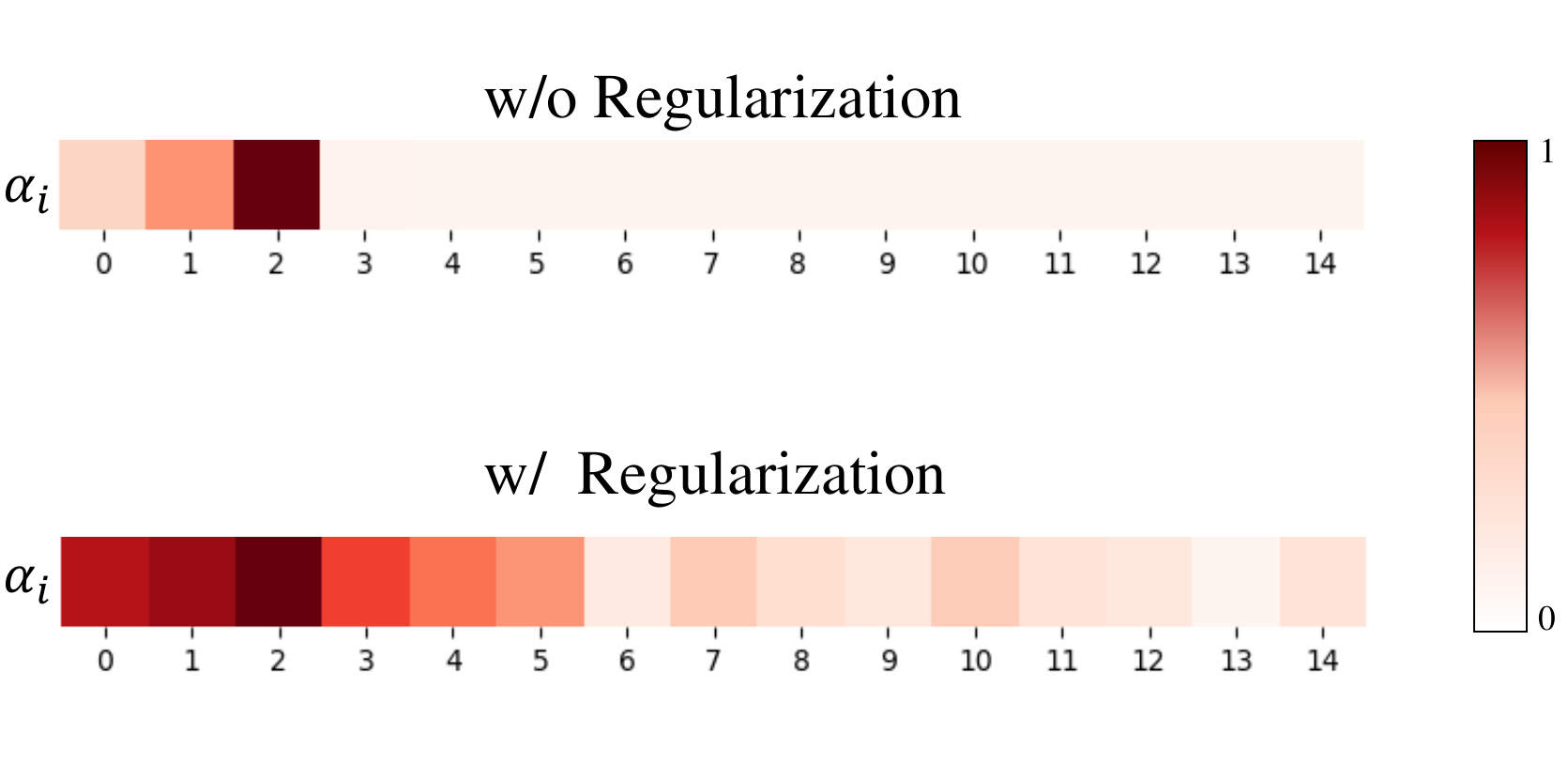}
        \caption{Visualization of attention weights $\alpha_i$ for two input samples $x_{1} \in \mathcal{X}$ and $x_{2} \in \mathcal{X}$
     with and without regularization. }
        \label{fig:awr_1}
    \end{subfigure}
    \hfill
    \begin{subfigure}{0.48\linewidth}
        \centering
        \includegraphics[width=1\linewidth]{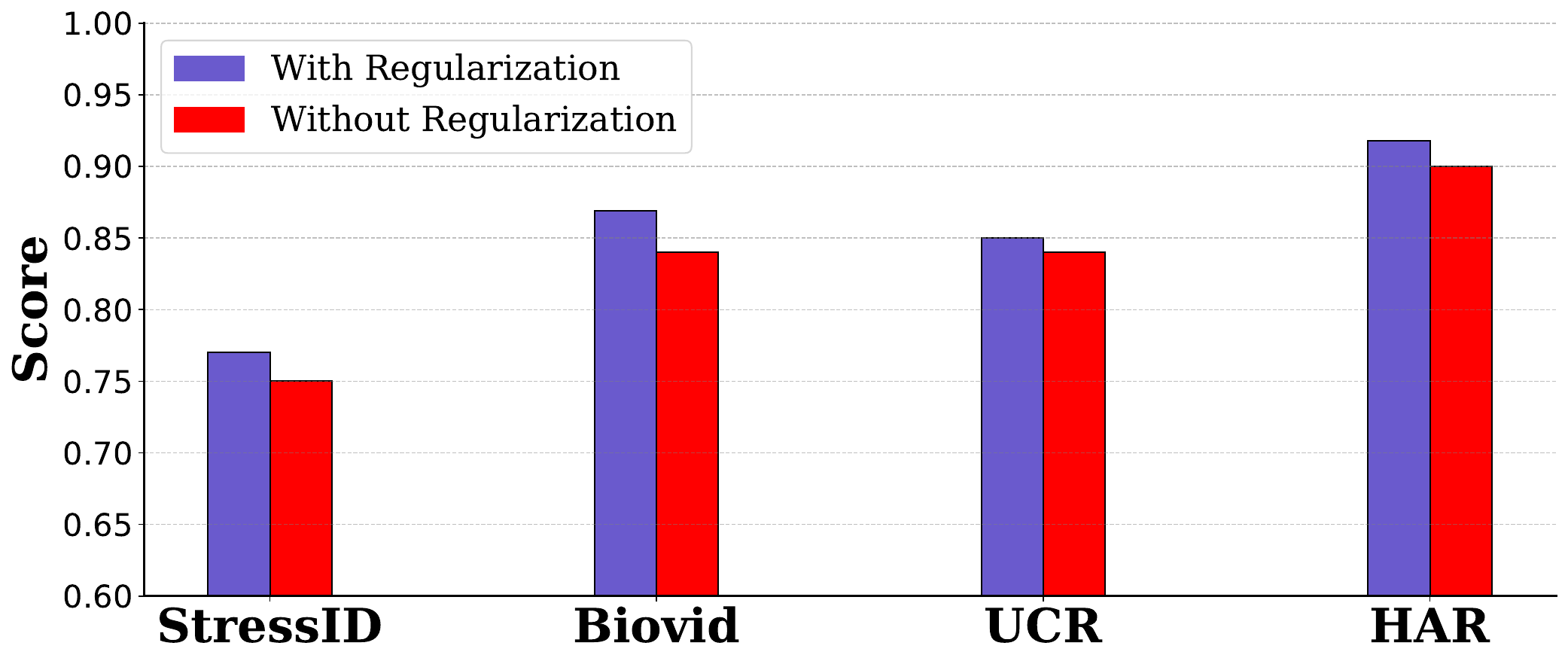}
        \caption{Performance comparison with and without attention weights $\alpha_i$ regularization on various datasets.}
        \label{fig:awr_2}
    \end{subfigure}
    \caption{Effect of attention weights $\alpha_i$ regularization on student performance}
    \label{fig:awr}
\end{figure}


\noindent\textbf{Effect of Attention Weights Regularization.}
SSD heavily relies on the diversity in the $\hat{f}^\mathcal{T}(x)$, implicitly mimicking the ensemble of task-relevant embeddings from the teacher space. If the attention weights are not regularized, the $\hat{f}^\mathcal{T}(x)$ would be highly influenced by one of the teacher embeddings, which is closest to the current $f^\mathcal{S}(x)$, essentially rendering the proposed methodology ineffective. Fig.~\ref{fig:awr_1} shows the attention weights of an input sample $x \in \mathcal{X}$. It can be observed from Fig.~\ref{fig:awr_1} that, without regularization, the attention weight is extremely high for $\alpha_{2}$. On the other hand, the weights are more spread out with regularization, showing that the student model trained with SSD leverages the diversity in the meaningful teacher representations. Fig. \ref{fig:awr_2} shows the results obtained with and without regularization of $\alpha_{i}$. In all instances, regularization improves accuracy.

\begin{figure*}[h]
    \centering
    \includegraphics[width=0.95\linewidth]{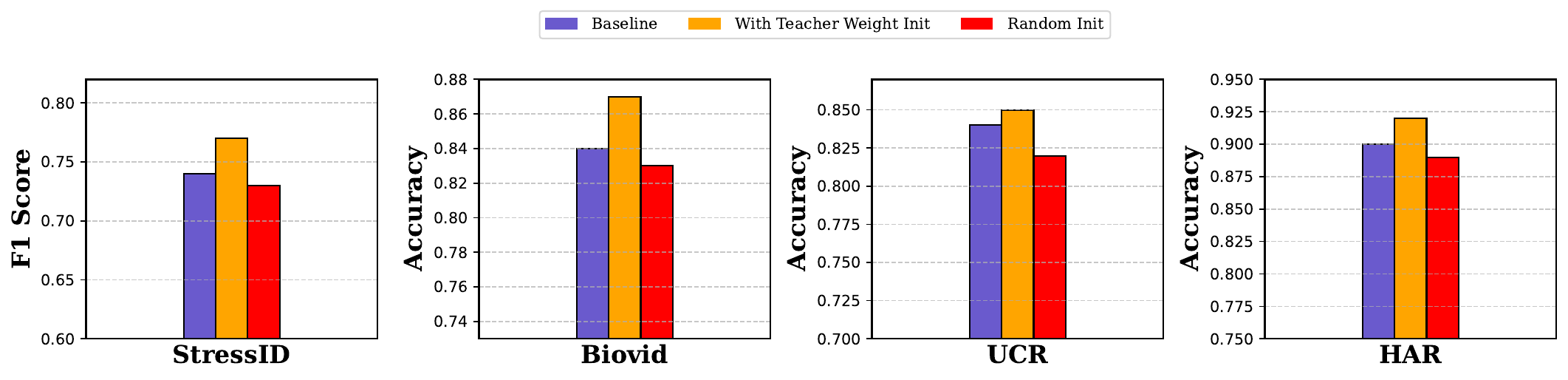}
    \caption{Performance of SSD with and without student parameters initialization with trained teacher weights $\theta^\mathcal{T'}$ on various datasets.}
    \label{fig:stu-weight}
\end{figure*}

\noindent\textbf{Effect of Student Parameters Initialization with Teacher Weights.}
\label{sub:stu_weight}
The current student representation guides the distillation process at each step. This section investigates the impact of student parameter initialization with the trained teacher weights $\theta^\mathcal{T'}$. Fig.~\ref{fig:stu-weight} shows the results obtained by the student model with the baseline, random initialization, and student parameter initialization with trained teacher weights. 
It can be observed from the figure that in each instance, the random initialization performs even worse than the baseline. This shows the effectiveness of the proposed method, because if the student model is not initialized with $\theta^\mathcal{T'}$, the $\phi_{i}$ calculated would be ineffective since the current $f^\mathcal{S}(x)$ is not task aligned. Hence adding the additional constraint in the $\mathcal{L}_{total}$ term breaks the performance. See Appendix C.1 for detailed results.

\subsection{Discussion}
\textbf{Analysis of Performance on Teacher Network Architecture.} SSD can improve performance quite significantly on real-world datasets, particularly with models that have multiple dropout layers built into the architecture. Methods like PAN \cite{Lu_pan2023} and the fusion-based architecture used for Biovid and StressID datasets, respectively, have multiple dropout layers throughout backbones and various modules; in contrast, models for HAR and CIFAR datasets are 1D and 2D-Conv ResNet-based architectures with only one dropout layer. One direct correlation between the performance and the capacity to generate stochastic representations can be drawn. The models with a higher number of dropout layers naturally have a greater capacity to generate stochastic representations with more diversity, which allows the proposed method to form a more informative $\hat{f}^{\mathcal{T}}(x)$, consequently leading to a student model with a significant performance boost over the teacher model.

\noindent\textbf{Supervised vs. Unsupervised Setting.} The proposed method can enhance performance both in supervised and unsupervised settings. The performance boost observed in the supervised setting is slightly higher than in the unsupervised setting; this could be because the unsupervised contrastive loss-based methods are already equipped with the ability to learn generalized representation. The augmented views of the input data in both TS2Vec and SoftCLT methods lead to more generalized representations. 

\noindent\textbf{Why is SSD Different from just Teaching \textit{Dropout} to the Student Model?} Intuitively, it seems that the proposed method might be an alternative way of teaching the student model to drop out. However, during both the teacher training step and student training, the standard \textit{training-time} dropout is activated, but the performance boost is not observed. The performance boost can be explained by the filtering of the teacher representations, where \textit{teaching the student to drop out} would be equivalent to distilling from all stochastic representations $f^\mathcal{T}_{i}(x)$ instead of the attended teacher representation $\hat{f}^\mathcal{T}(x)$, which is then filtered through the proposed SSD method (see Appendix D for a detailed discussion). This explanation is also supported by the results in Table \ref{tab:abb-nr} where we first distill from all $f^\mathcal{T}_{i}(x)$, and the student performance does not improve.

\section{Conclusion}
In this work, we introduced SSD, a novel approach to enhancing diversity in the teacher space by leveraging the stochastic nature of DL models  using \textit{distillation-time} dropout and applying SGKD to learn meaningful representations.  It employs the student's current representation as a guide to select meaningful representations, implicitly mimicking an ensemble of task-relevant representations. 
Extensive experiments on real-world time-series data, complemented by validation on a benchmark vision dataset, show the effectiveness of SSD in improving representation learning. While SSD outperforms SOTA methods, its current evaluation is limited to architectures that already incorporate dropout. Given that SSD operates in the latent space, we hypothesize that the proposed SGKD framework could be extended to other settings where diversity is introduced through perturbations in the feature space or noise injection. This presents an compelling avenue for future research and a promising alternative to deep ensembles. It also provides an alternative for learning generalized representations for time-series data, paving the way for more efficient and robust representation learning in a wide range of applications.

\begin{credits}
\subsubsection{\ackname} 
This research endeavor was partially supported by the Natural Sciences and Engineering Research Council of Canada (NSERC), Fonds de recherche du Québec – Santé (FRQS), Canada Foundation for Innovation (CFI), and the Digital Research Alliance of Canada.

\subsubsection{\discintname}
The authors have no competing interests to declare that are relevant to the content of this article.
 \end{credits}
%
%
%
\bibliographystyle{splncs04}
%

\newpage
\appendix

\section{Datasets and Implementation Details}

\subsection{Datasets}
\label{app:datsets}
\noindent\textbf{Biovid Heat Pain Database:} The Biovid Heat Pain Database is among the most popular databases for pain estimation. The dataset is divided into five parts. We use part A and part B of the dataset. Part A of the dataset contains videos and physiological signals, including GSR, EMG, and ECG. The dataset is annotated for discrete labels for pain intensities, where BL refers to 'baseline/no pain' and PA1-PA4 refers to increasing pain intensities. The Biovid part A has 87 subjects, each with 100 videos, corresponding to 20 videos per class, which results in a total of 8700 videos. Twenty subjects did not exhibit any noticeable response to the pain stimulus, so some studies report results for the remaining 67 subjects. To validate the proposed method, we report results on the entire dataset of 87 subjects. In addition to the modalities available in part A, part B of the dataset also has facial EMG and is recorded for 86 subjects, corresponding to 8600 videos. The Biovid dataset does not come with predefined training, validation, or test splits, so many studies have validated their methods using cross-validation. Following that, we also validate the proposed method using leave-one-subject-out (LOSO) cross-validation (CV). The performance metric used is accuracy.
\\
\\
\noindent\textbf{StressID Dataset:} The StressID \cite{stressid2023} dataset is a multimodal dataset for stress identification. It contains physiological signals, including electrocardiography (ECG), electrodermal activity (EDA), respiratory rate (RR), audio recordings, and facial recordings. Cognitive tasks like public speaking scenarios, comprehension, or mathematical exercises are used as stress-inducing stimuli. The total number of participants is 65, performing 11 tasks. The dataset is annotated with both binary classes, i.e., stressed vs. non-stressed, and three classes, i.e., relaxed, neutral, and stressed. Both accuracy and F1 are used as performance metrics. We follow the same evaluation protocol as the original dataset authors \cite{stressid2023} and report both accuracy and weighted F1 score on binary and 3-class problems. 
\\
\\
\noindent\textbf{Time-Series Benchmark Datasets:} The UCR Archive \cite{dau2019ucrtimeseriesarchive} serves as a benchmark repository for researchers working on time-series classification, clustering, and related tasks. We present results on the time-series classification task on the wearable and biosignal datasets from the UCR Archive. Predefined train and test splits are provided for each dataset and the performance metric used for all datasets is accuracy. 
\\
\\
\noindent\textbf{Human Activity Recognition (HAR) Dataset:} The UCI HAR dataset \cite{Har-ds} is a collection of waist-mounted smartphone-based sensor readings for 30 subjects. The signals are recorded while performing six activities, i.e., lying down, walking, walking upstairs, downstairs,  sitting, and standing. The data is collected using the accelerometer and gyroscope at a sampling rate of 50 Hz.

\noindent\textbf{CIFAR10 and CIFAR100:} The CIFAR-10 and CIFAR-100 datasets are widely used benchmarks in the field of computer vision. Both datasets were created by the Canadian Institute for Advanced Research (CIFAR) and contain 60,000 32$\times$32 color images. The CIFAR-10 dataset is composed of 10 mutually exclusive classes, each containing 6,000 images, representing a diverse set of objects such as airplanes, automobiles, birds, and cats. CIFAR-100 dataset extends this diversity by including 100 different classes, with 600 images per class, providing a more challenging problem due to the larger number of categories and increased intra-class variance.

\subsection{Network Architecture for the HAR Dataset}
\label{app:har_arch}
The 1D CNN used for the HAR dataset is a custom CNN with two convolutional layers and three fully connected layers. Table \ref{tab:network_architecture} shows the network architecture.

\begin{table}[t]
    \centering
        \caption{1D CNN architecture used with the HAR dataset.}
    \begin{tabular}{c|c|c|c}
        \hline
        \textbf{Layer} & \textbf{Type} & \textbf{Kernel / Features} & \textbf{Activation} \\
        \hline \hline
        Conv1 & Conv2d & $(9, 32), \text{kernel}=(1,9)$ & ReLU \\
        & MaxPool2d & $(1,2), \text{stride}=2$ & - \\
        \hline
        Conv2 & Conv2d & $(32, 64), \text{kernel}=(1,9)$ & ReLU \\
        & MaxPool2d & $(1,2), \text{stride}=2$ & - \\
        \hline
        FC1 & Linear & $1664 \rightarrow 1000$ & ReLU \\
        \hline
        FC2 & Linear & $1000 \rightarrow 500$ & ReLU \\
        \hline
        FC3 & Linear & $500 \rightarrow 6$ & - \\
        \hline
    \end{tabular}
    \label{tab:network_architecture}
\end{table}

\noindent\textbf{Implementation Details - HAR Dataset: }
For validation on Human Activity Recognition (HAR) Dataset, we use 1D CNN barebones as shown in Table \ref{tab:network_architecture}. The teacher model was trained for a total of 100 epochs with Adam optimizer. The initial learning rate was 0.05 with a \textit{ReduceLROnPlateau} scheduler on training loss with patience of 10 and a reduction factor of 0.1. While training the student, the total number of repetitions to generate teacher representations was 30. The batch size was 128, and the distillation-time dropout rate was 0.2.

\subsection{Validating SSD Beyond Time-Series}
\label{app:subsec_bts}The proposed method is mainly validated on time-series signals of two real-world affective computing datasets and wearable time-series datasets from the UCR Archive. Since the SSD relies on diversification in feature space, the same concept can be extended to vision datasets. To prove the effectiveness of the proposed method across a variety of tasks, we perform additional validation on benchmark vision datasets CIFAR-10 and CIFAR-100. The proposed method achieves $94.83\%$, improving $1.5\%$ over the baseline with standard ResNet34 w/dropout. For CIFAR100, the model improves $1.2\%$ over the baseline using ResNet50. By applying the proposed SSD method, we achieve $79.47\%$, which is improved after distillation to $81.73\%$.

\begin{table}[!h]
\centering
\small
\caption{Additional evaluation on the CIFAR-10 and CIFAR-100 datasets.}
\begin{tabular}{c|c|c|c}
\hline
\textbf{Method}              & \textbf{Model}                     & \textbf{Dataset}                   & \textbf{Accuracy (\%)}       \\  \hline \hline
Baseline (\textit{w/o} Distillation)    & \multirow{2}{*}{ResNet34} & \multirow{2}{*}{CIFAR-10}  & 93.23          \\
\textbf{SSD (ours)} &                           &                           & \textbf{94.83} \\ \hline
Baseline (\textit{w/o} Distillation)    & \multirow{2}{*}{ResNet50} & \multirow{2}{*}{CIFAR-100} & 79.47          \\
\textbf{SSD (ours)} &                           &                           & \textbf{81.73}   \\ \hline
\end{tabular}

\label{tab:abb-cifar}
\end{table}

\noindent\textbf{Implementation Details - CIFAR-10 and CIFAR-100.} For validation on vision datasets, we use the barebones ResNet34 for CIFAR-10 and ResNet50 for CIFAR-100. The teacher model for CIFAR-10 was trained for a total of 120 epochs with an SGD optimizer. The initial learning rate was 0.1 with a \textit{ReduceLROnPlateau} scheduler on training loss with patience of 10 and a reduction factor of 0.1. The total number of repetitions to generate teacher representations was 30. The batch size was 125, and the distillation-time dropout rate was 0.2. For CIFAR-100, we use ResNet50 as the backbone. The model was trained for a total of 120 epochs. The network was optimized using the SGD optimized with the initial learning rate of 0.1 and reduced with \textit{cosineAnnealingLR} scheduler. Total teacher repetitions $n$ were 30, and the dropout rate $p$ was 0.2.

\section{Comparison of Train Time Complexity}
\label{app:train_rep_comparison}

This section compares the training time and computational complexity of SSD with traditional ensembles, Stochastic Weight Averaging, and Model Soups. SSD performance at test is at par with state of the art Greedy Soup method, however in terms of train-time computational complexity, the proposed method only adds insignificant overhead. For methods like greedy soup, the model needs to be trained 25$\times$ whereas in SSD, the model is trained only two times.
\begin{figure}[!h]
    \centering
    \includegraphics[width=1.05\linewidth]{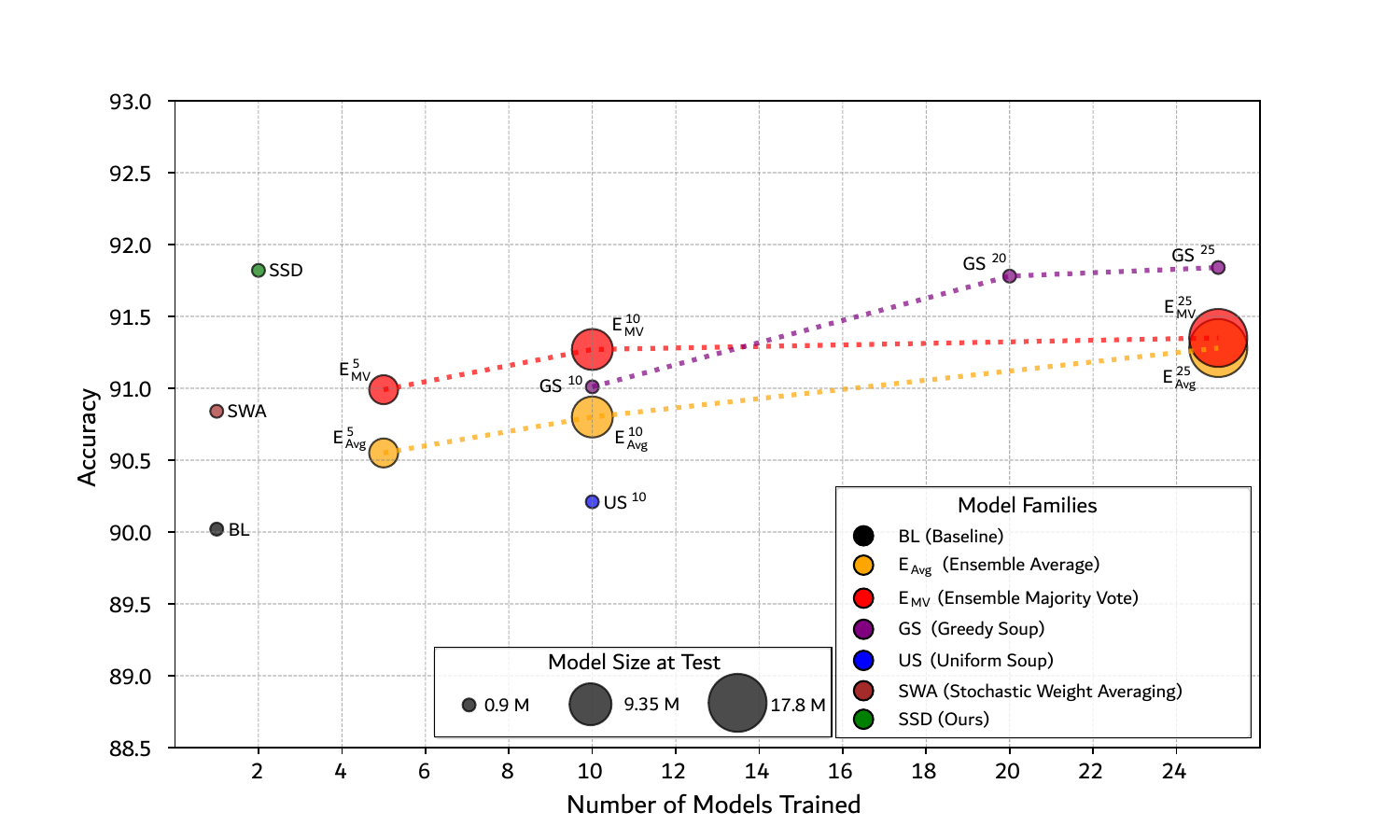}
\caption{Accuracy on HAR data of the SSD method with baseline (BL), traditional ensembles with majority vote ($\text{E}_{\text{MV}}$) and average ($\text{E}_\text{Avg}$), uniform soup (US) and greed soups (GS) in terms of model size at inference and training time computation.}
    \label{fig:comp-perf-trainrep}
\end{figure}

\section{Additional Results}
\label{sec:add_abl}

\subsection{Effect of Student Parameters Initialization with Teacher Weights}
\label{app:stu_weight}
\begin{table}[!h]
\centering
\caption{Performance of the proposed SSD method with 1D-CNN on the HAR dataset with and without initialization with teacher weights $\theta^\mathcal{T}$.}
\begin{tabular}{c|c}
\hline
\textbf{Model}  & \textbf{Accuracy} \\ \hline \hline
1D CNN (Teacher) & 0.9002   \\ \hline
Student w/o $\theta^\mathcal{T}$   & 0.8975   \\ \hline
\textbf{SSD (ours)}    & \textbf{0.9182}   \\ \hline
\end{tabular}

\label{app:stw-har}
\end{table}

The current student representation is what guides the distillation process at each distillation step. This section investigates the impact to student parameter initialization with the trained teacher weights $\theta^\mathcal{T'}$. Table \ref{app:stw-har} compares the performance of SSD on the HAR dataset. The baseline i.e. the teacher model achieves $0.9002$ accuracy score. However the student model performance with random initialization is 0.8975, which is lower than the baseline performance. As explained in Section 4.3, this decrease in performance is observed because the distillation process is guided by the current student representation $f^{\mathcal{S}}$. If the student model is not initialized with the teacher weights, it learns a separate trajectory to converge. This phenomenon is similar to the concept of pretraining in weight averaging methods, where if weights of different models trained from scratch are averaged, the performance diminishes drastically because each model learns a different trajectory towards convergence. On the other hand, if the models start with pre-trained weights, they are more likely to converge with similar weights, and averaging them leads to better performance. Similarly, in SSD, if the student model parameters are randomly initialized, the student representation $f^{\mathcal{S}}$ for each $x \in X$  will be misaligned with the teacher representation. It is only after the student model parameters with $\theta^\mathcal{T}$, that the $\phi_i$ calculated is aligned leading to a teacher representation $\hat{f}^\mathcal{T}$ that has meaningful diversity aligned with the student representation.   

\begin{table*}[h]
\centering
\caption{Accuracy of UCR datasets with and without student parameters initialization with $\theta^{\mathcal{T}}$ }
\small
\begin{tabular}{l|>{\centering\arraybackslash}m{2.5cm}|c}
\hline
\textbf{Dataset} & \textbf{SSD} & \textbf{Student w/o $\theta^{\mathcal{T}}$} \\ \hline \hline
ECG200                     & 0.9300 & 0.8478      \\
ECG5000                    & 0.9413 & 0.9187      \\
TwoLeadECG                 & 0.9798 & 0.9625      \\
NIFetalECGThorax1          & 0.9394 & 0.9112      \\
NIFetalECGThorax2          & 0.9480 & 0.9326      \\
Chinatown                  & 0.9708 & 0.9618      \\
UWaveGestureLibraryX       & 0.8143 & 0.7829      \\
UWaveGestureLibraryY       & 0.7266 & 0.6924      \\
UWaveGestureLibraryZ       & 0.7682 & 0.7518      \\
MedicalImages              & 0.7710 & 0.8017      \\
DodgerLoopDay              & 0.5500 & 0.5032      \\
DodgerLoopGame             & 0.8695 & 0.7914      \\ \hline
Total                      & 0.8508 & 0.8215      \\ \hline
\end{tabular}
\label{app_table:stw-wearablebio}
\end{table*}

Table \ref{app_table:stw-wearablebio} also presents similar results where the average of the 12 selected datasets from the UCR archive, drops from 0.8426 to 0.8125 with random initialization of student parameters. Whereas the student model with teacher weights initialization achieves 0.8508 average accuracy score.

\subsection{Effect of Percentile Thresholding on Performance}
The proposed SSD method relies on the dynamic selection of the teacher representation based on percentile thresholding. The parameter $\epsilon$ controls the percentile value. This section studies the effect of different $\epsilon$ values on the overall student performance.

\begin{table}[!h]
\centering
\caption{Effect of $\epsilon$ on the student performance}
\begin{tabular}{c@{\hskip 10pt}|@{\hskip 10pt}c@{\hskip 10pt}c@{\hskip 10pt}c@{\hskip 10pt}c@{\hskip 10pt}c@{\hskip 10pt}c@{\hskip 10pt}c@{\hskip 10pt}c}
\hline
\textbf{$\epsilon$} & 20 & 30 & 50 & 60 & 70 & 80 & \textbf{90} & 100 \\ \hline
\textbf{Accuracy} & 0.9024 & 0.9082 & 0.9082 & 0.9085 & 0.9152 & 0.9152 & \textbf{0.9182} & 0.9148 \\
\hline
\end{tabular}
\label{tab:abb-epsilon-har}
\end{table}

Table \ref{tab:abb-epsilon-har} shows the student performance with increasing value of $\epsilon$. For lower values of $\epsilon$, the student performance is close to the baseline. As $\epsilon$ increases, the far-off representations in the teacher space are effectively masked/filtered, allowing only relevant ones to be included in the $\hat{\alpha}_i$. The model's performance peaks at $\epsilon=90$. A decrease in performance is observed again at $\epsilon=100$, likely because this high threshold selects only the single teacher representation most similar to the student representation, significantly limiting diversity in the teacher and the student representations. Another interesting observation is that the performance is not as sensitive to the $\epsilon$ value as it is to statically selecting top-$k$ representations. This is because the top-$k$ is unaware of batch dynamics always selects a $k$ number of representations, whereas $\epsilon$-thresholding dynamically selects only the relevant teacher representations.


\section{Why \textit{SSD} is different from simply \textit{Teaching Student to Dropout}?}
\label{app:th}
Let us take the following scenarios where 1) teaching the student to dropout, and 2) SSD. Let $\{f_i^\mathcal{T}(x)\}_{i=1}^n$ be teacher outputs (representations) with dropout, and $\hat{f}^\mathcal{T}(x)=\mathcal{A}(\{f_i^\mathcal{T}(x)\})$ be the attended teacher representation via student guided attention $\mathcal{A}$. Let $g_S(x; D)$ be the student output (representation) under dropout mask $D$,
and $\mathbb{E}_D[\cdot]$ be the expected representation of the student over dropout masks. Consider the following two losses : 
$\mathcal{L}_{\text{dropout}} = \big[ \| g_\mathcal{S}(x; D) - \mathbb{E}_D[f^\mathcal{T}(x)] \|^2 \big]$ and
$\mathcal{L}_{\text{SSD}} = \| g_\mathcal{S}(x) -\mathbb{E}_D[ \hat{f}^\mathcal{T}(x)] \|^2$. 
\\
$\mathcal{L}_{\text{dropout}} = \mathcal{L}_{\text{SSD}}  \text{ if and only if: } \mathbb{E}_D[f^\mathcal{T}(x)] = \mathcal{A}(\{f_i^\mathcal{T}(x)\}).$ This equality does not hold.



\noindent\textbf{Contradiction:} $\mathbb{E}_D[f^\mathcal{T}(x)]$ is the unfiltered mean of $\mathcal{f}_i^\mathcal{T}(x)$, including noise. Whereas,
 $\hat{f}^\mathcal{T}(x) = \mathcal{A}(\{f_i^\mathcal{T}(x)\})$ selectively filters $\{f_i^\mathcal{T}(x)\}$, removing noise and provides an attention weighted teacher representation. 
 \\
 Thus, $\mathbb{E}_D[f^\mathcal{T}(x)] \neq \mathcal{A}(\{f_i^\mathcal{T}(x)\})$, contradicts the assumption. $\mathcal{L}_{\text{SSD}}$ ensures $g_\mathcal{S}(x)$ aligns with $\mathcal{A}(\{f_i^\mathcal{T}(x)\})$, improving task relevance. $\mathcal{L}_{\text{dropout}}$ aligns $g_\mathcal{S}(x; D)$ with $\mathbb{E}_D[f^\mathcal{T}(x)]$, which includes noise, reducing task specificity. $\mathcal{L}_{\text{SSD}}$ achieves superior task performance by selective attention to relevant teacher outputs.

\bibliographystyle{splncs04}
%

\end{document}